\documentclass[conference]{IEEEtran}
\IEEEoverridecommandlockouts 
\usepackage[numbers]{natbib}
\usepackage{amsmath,amssymb,amsfonts}
\usepackage[ruled,vlined]{algorithm2e}
\usepackage{graphicx}
\usepackage{textcomp}
\usepackage{hyperref}
\usepackage{color,xcolor}
\usepackage{booktabs}
\definecolor{linkcolor}{rgb}{1.0,0.078,0.576}
\hypersetup{
	colorlinks=true,
	allcolors=linkcolor,
}

\ifCLASSOPTIONcompsoc
\usepackage[caption=false,font=normalsize,labelfon
t=sf,textfont=sf]{subfig}
\else
\usepackage[caption=false,font=footnotesize]{subfig}
\fi

\begin{document}

	\title{
		ARC: A Vision-based Automatic Retail Checkout System
	\thanks{
		This work was completed as a Senior Year Project in the 2017-18 academic year, funded by the Department of Mechatronics and Control Engineering.
		The manuscript has previously been rejected at multiple conferences.
		The authors graduated and went their own ways in life, and therefore, this work will be now be carried forward by others.
		Dataset of retail items and source code can be accessed at \texttt{\url{https://github.com/stalhabukhari/ARC}}.
		}
	}

	\author{
		\IEEEauthorblockN{Syed Talha Bukhari, Abdul Wahab Amin, Muhammad Abdullah Naveed, Muhammad Rzi Abbas\IEEEauthorrefmark{1}}
		\IEEEauthorblockA{
			\textit{Department of Mechatronics and Control Engineering} \\
			\textit{University of Engineering and Technology, Lahore (Main Campus)}, Pakistan
		}
		\IEEEauthorrefmark{1}Email: \href{mailto:muhammadrziabbas@uet.edu.pk}{\texttt{muhammadrziabbas@uet.edu.pk}}
	}

	\maketitle

	\begin{abstract}
		Retail checkout systems employed at supermarkets primarily rely on barcode scanners, with some utilizing QR codes, to identify the items being purchased.
		These methods are time-consuming in practice, require a certain level of human supervision, and involve waiting in long queues.
		In this regard, we propose a system, that we call \emph{ARC}, which aims at making the process of check-out at retail store counters faster, autonomous, and more convenient, while reducing dependency on a human operator.
		The approach makes use of a computer vision-based system, with a Convolutional Neural Network at its core, which scans objects placed beneath a webcam for identification.
		To evaluate the proposed system, we curated an image dataset of one-hundred local retail items of various categories.
		Within the given assumptions and considerations, the system achieves a reasonable test-time accuracy, pointing towards an ambitious future for the proposed setup.
		The project code and the dataset are made publicly available.
	\end{abstract}


	\section{Introduction}\label{S1}
	\IEEEPARstart{S}{}upermarkets have an enormous catalogue of products of different varieties, and as innovations keep burgeoning, this number continues to grow.
	The availability of a huge number of diverse products invites the implementation of advanced and autonomous techniques to manage the database from both ends, i.e. when something is either added or removed.
	The removal process typically comprises of selling a product to a customer, where a retail checkout platform employing a bar-code scanner is used.

	Normally, each product is screened through a barcode test by the operator who manually handles the product, locates the barcode, and hovers the scanner over it.
	This time-consuming task creates a bottleneck in the shopping experience, especially at large supermarkets, where customers have to wait in long queues to check-out.
	To address this problem, new methods aiming at enhancing the shopping experience and making it seamless are being devised.
	For example, Amazon Go~\cite{ref_amazon_go} aims at revolutionizing the shopping experience by removing the complete checkout counter altogether.

	The research-active field of artificial intelligence is making breakthroughs with developments in deep learning~\cite{ref_DL} over the past decade, and the immense potential is yet to be explored.
	These algorithms require less pre-processing and perform feature extraction efficiently on their own without explicit instruction.
	This independence from explicit modeling and human intervention makes the process of implementing them less onerous, even more so with the availability of open-source deep learning frameworks such as Keras~\cite{ref_keras}.
	A major bottleneck, however, is the requirement of large computational budgets to train deep neural networks and to translate them to portable hardware~\cite{ref_spikingnns,ref_quantnns}.

	We make use of deep learning in our proposed setup that we call ARC (\textbf{A}utomatic \textbf{R}etail \textbf{C}heckout) in the form of a light-weight Convolutional Neural Network which identifies retail items from their images taken by a webcam.
	The motive behind the effort is that humans make use of visual input \textit{alone} to identify retail items, and therefore, a reliable visual feedback mechanism in machines can obviate the need for alternatives such as barcode scanners.

	\section{Related Work}\label{S1_2}
	The recent success of Amazon Go~\cite{ref_amazon_go} has provoked great interest in self-checkouts at grocery stores.
	Amazon Go stores have eliminated the dependency on conventional checkout counters and significantly reduced the time for checkouts.
	The method involves use of computer vision and sensor fusion to classify which products have been purchased, subsequently charging the customers automatically through the a mobile application when they exit the store.
	This removes the inconvenience of long queues, hence significantly reducing check-out time.
	Panasonic has introduced a walk-through self-checkout system~\cite{ref_panasonic} based on radio frequency identification (RFID) tags.
	Detection of objects through RFID has been implemented before~\cite{ref_rfid_patent}, the major contribution here is a cost-effective solution for deployment at grocery stores.

	The problem of visual object recognition and classification has been extensively studied by the research community and work has been done on product detection and classification in grocery stores, especially on detection of products on shelves.
	In \cite{ref_mobile}, low level features are extracted from the query image using CHoG descriptor and sent to a data server for recognition.
	The proposed approach focuses more on application development, with relatively little attention to the recognition problem.
	\cite{ref_grozi} introduces a dataset comprising $120$ grocery products (GroZi-120) comprising pictures of products taken from the web (in-vitro) as well as extracted from a camcorder video of retail items recorded inside a grocery store (in-situ).
	Authors utilized color histogram, SIFT and boosted Haar-like features on the dataset and a comparison against product type and imaging conditions is done.
	\cite{ref_surf} uses enhanced SURF descriptors extracted from the GroZi-120 dataset for training a multi-class Naive-Bayes classifier.
	The authors propose a system called \emph{ShelfScanner} aiming to help visually impaired people to shop at grocery stores.

	Closest to our approach is \cite{ref_ISCOS} which proposes an \emph{Intelligent Self-Checkout System} (ISCOS) using a single camera to detect multiple products in real-time.
	Using web scraping, product data is collected from webpages of three markets and images are gathered from three image search engines for training of a YOLO classifier~\cite{ref_yolo}.
	The arduous task of manual annotation is simplified by applying background subtraction to bound the locations of products automatically.
	The system identifies multiple objects in the camera's field-of-view over a non-moving platform, until the number goes beyond three where it struggles to detect small objects in the presence of larger objects.
	Furthermore, the system only makes use of coarse features and does not delve into the finer details of each item.

	Different from the approaches mentioned above, we propose a cost-effective checkout system based on a motorized conveyor mechanism, which carries each retail item into the system.
	Appearance of the item is captured using a webcam and processed before sending to a convolutional neural network for identification.
	In our setup, we treat the object identification task as an $N$-class classification problem.
	Remainder of this paper is organized as follows:
	Section~\ref{sec:config} describes the system configuration, Section~\ref{sec:method} details the underlying Methodology, Section~\ref{sec:experiment} discusses implementation and results, and Section~\ref{sec:conclude} summarizes the effort.

	\begin{figure}[!t] 
		\centering
		\includegraphics[width=2.5in]{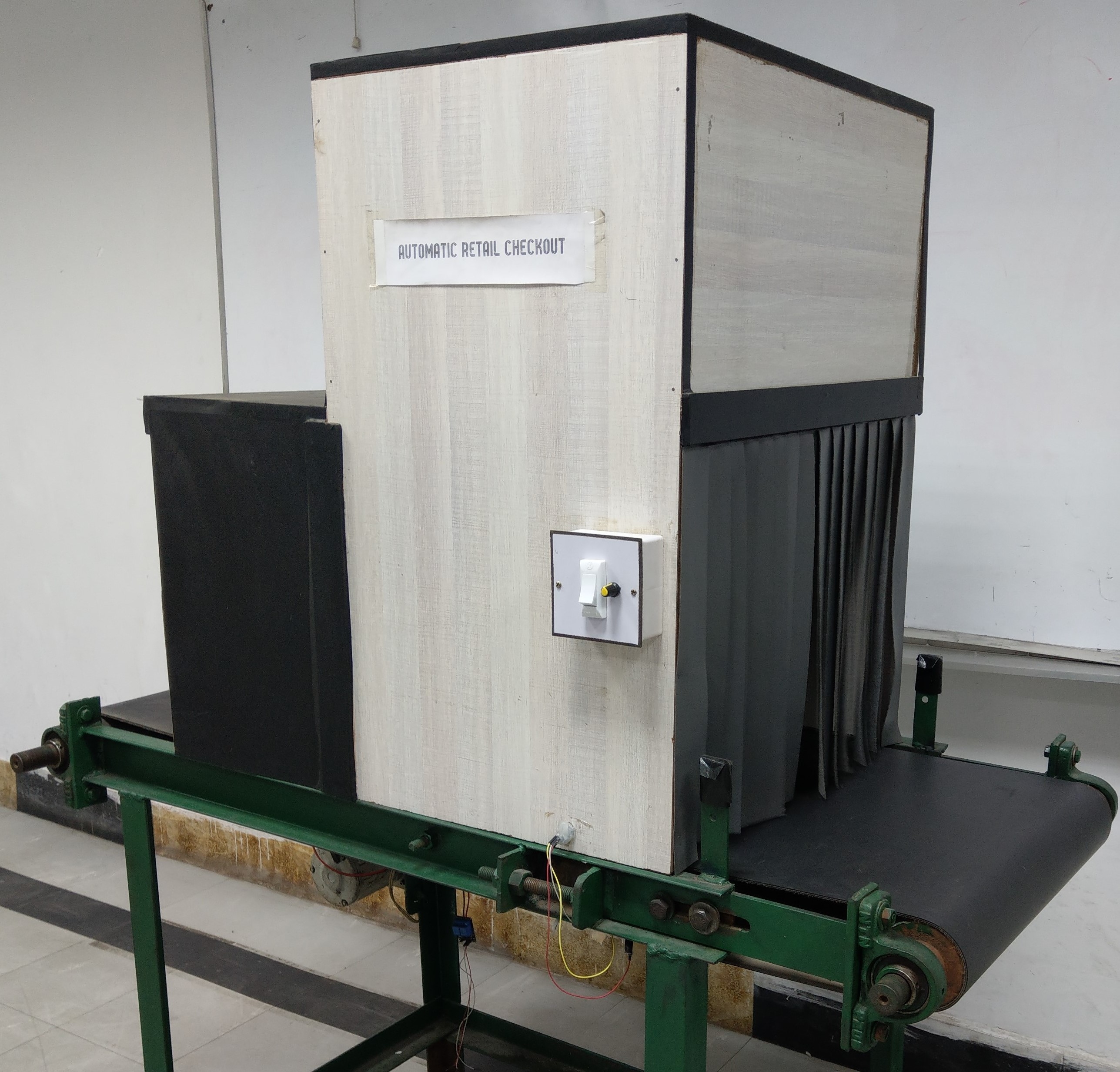}
		\caption{
			\textbf{Design Prototype}.
			A designed prototype of the proposed checkout system is shown.
			It consists of a motor-powered conveyor-belt mechanism with a wooden hood attached over it.
			The interior of the hood is illuminated by LED strips and contains a webcam that observes retail items fed to it.
		}
		\label{fig:design_proto}
	\end{figure}

	\section{System Configuration}\label{sec:config}

	\subsection{Hardware Setup}

	\begin{figure}[!t]
		\centering
		\includegraphics[width=3in]{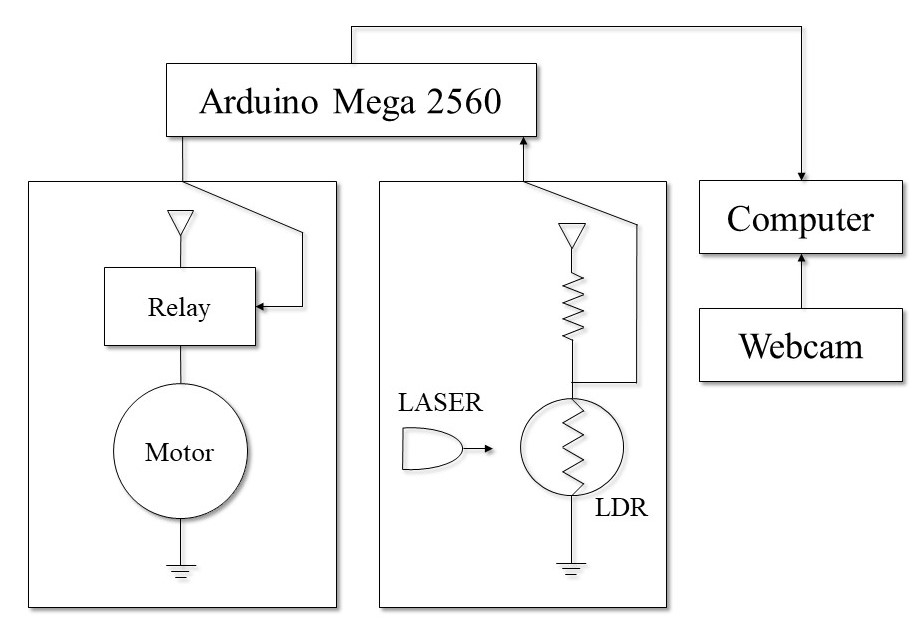}
		\caption{
			\textbf{Hardware Setup Overview}.
			An Arduino Mega 2560 Microcontroller is interfaced with the computer through serial communication, and controls motor switching (left box) in accordance with the signal received from LASER-LDR module (middle box).
			The computer receives continuous feed from the webcam and decides when to extract a frame based on signals from the microcontroller.
		}
		\label{fig:setup_overview}
	\end{figure}

	The designed setup (Fig.~\ref{fig:design_proto}) sits over a conveyor-belt mechanism (belt length: $9$’, belt width: $1.5$’, belt thickness: $3\mathrm{mm}$) powered by a single-phase induction motor.
	A wooden hood houses a Logitech C310 webcam at the top, surrounded by an arrangement of $12\mathrm{V}$ LED strips to illuminate the interior (hood length: $29$”, hood width: $23$”, hood heights: $32$” for the LED and webcam housing, $18$” for the product receiving end).
	An Arduino Mega 2560 microcontroller controls motor switching through A $5$V relay module, depending on the input from a LASER-LDR (Light Dependent Resistor) module, and accordingly communicates with the Python environment on the computer which accesses the webcam for a frame from a continuous feed.
	The entrance to and exit from the wooden hood are covered by shutter strips cut from plastic table cloth, which (along with the LEDs) provide a closed and consistent environment inside the hood.
	Luminance of the LEDs was set to approximately $70\mathrm{lx}$ which we found adequate for a sufficiently clear image through our webcam.
	Additionally, an hp LaserJet printer is connected to the computer to print out receipts when all items purchased by a customer have passed.

	\subsection{Software Setup}
	The software setup is based on Python with OpenCV~\cite{ref_opencv} for majority of the image processing and Keras~\cite{ref_keras} (using TensorFlow backend) for implementing the object identification model (neural network).
	Python environment in the computer is interfaced with the Arduino microcontroller through serial communication via pySerial~\cite{ref_pyserial}.
	A GUI of the system is developed with Tkinter~\cite{ref_tkinter}.

	\begin{algorithm}
	\caption{Operating Procedure}\label{algo:procedure}

	begin customer bill, start conveyor motor.\\

	\While{cart not empty}{

		\underline{\textbf{input:}}\\
		feed item to conveyor.\\
		\underline{\textbf{detection:}}\\
		\If{LASER path is interrupted}{
			stop conveyor motor.\\
			extract webcam frame, pre-process, and send to neural network.\\
			start conveyor motor.\\
		}
		\underline{\textbf{output:}}\\
		\If{object identified}{
			update customer bill.\\
		}
	}
	generate final bill.\\
	\end{algorithm}

	\section{Methodology}\label{sec:method}
	The complete operating procedure is summarized in Algorithm~\ref{algo:procedure}.
	The operating procedure is divided into 4 major steps:
	Image Acquisition, Pre-processing, Object Identification, and Billing.

	\subsection{Image Acquisition}
	A retail item of adequate size is placed on one end of the running conveyor as it enters the hood, pushing the shutters aside.
	Inside the hood, as soon as the item interferes with the path of LASER-LDR module (indicating the arrival of the object within the field-of-view of webcam), the microcontroller switches off the conveyor motor.
	Simultaneously, a signal is sent to the Python environment to extract a frame from the continuous webcam feed, which is provided as input to subsequent processes.
	Once the image has been acquired, the conveyor motor restarts so that the item exits the hood.
	We do this to avoid motion blur in the extracted webcam frame.

	\subsection{Pre-processing}\label{ssec:preprocessing}
	\begin{figure*}[!t]
		\centering
		\subfloat[Webcam-acquired image (size: $640\times480$)]{\includegraphics[width=1.6in]{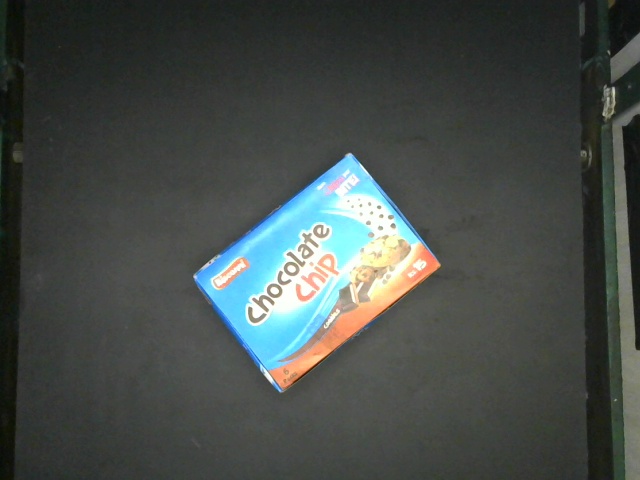}
			\label{subfig:preprocessing_1}}
		\hfil
		\subfloat[Our pre-processing pipeline.]{\includegraphics[width=2.1in]{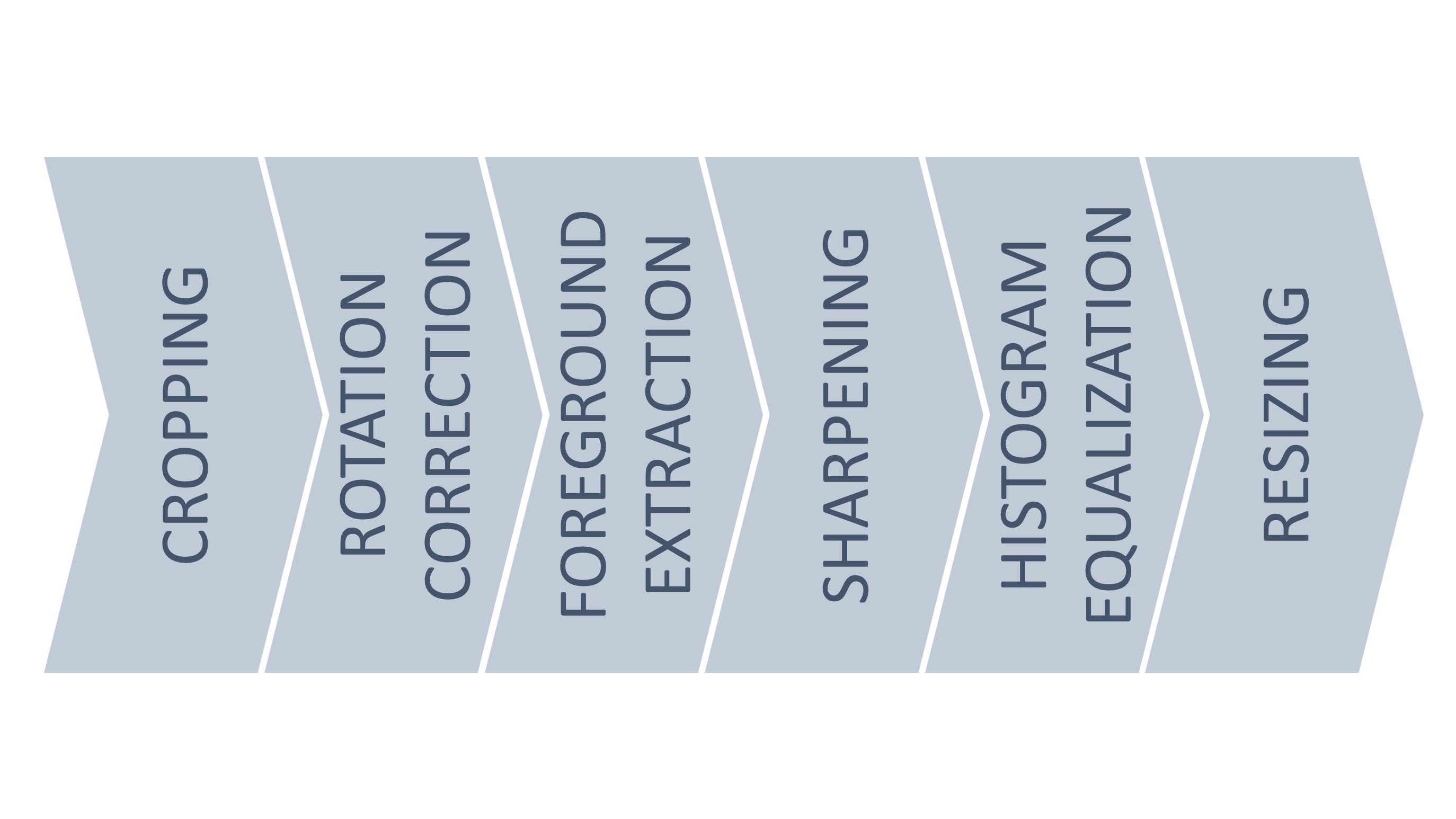}
			\label{subfig:preprocessing_steps}}
		\hfil
		\subfloat[Pre-processed image (size: $150\times150$)]{\includegraphics[width=1.2in]{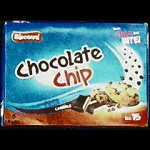}
			\label{subfig:preprocessing_2}}
		\caption{
			\textbf{Pre-processing Pipeline}.
			A sample input to and output from the pre-processing pipeline is shown.
			To preserve aspect ratio of the retail item during the resizing operation, the image is first zero-padded (hence the empty/black portions around the item after pre-processing).
			The output from this pipeline is fed to the deep neural network.
		}
		\label{fig:preprocessing}
	\end{figure*}

	The captured image passes through six pre-processing steps.
	First, portions of image containing unnecessary edges outside the conveyor belt area are cropped out (deterministic cropping, tuned manually).
	A binary convex hull of the resulting image is obtained and object orientation information is extracted via the spatial distribution of foreground pixels.
	This information is leveraged to undo the object's rotation.
	Although this stage may be bypassed, standardizing the orientation of each retail item lowers the burden on the object identification model, allowing it to focus more on the features characteristic to the retail item.
	This is further motivated by the fact that the convolution operation in Convolutional Neural Networks (our object identification model of choice, described in Section~\ref{ssec:cnn}) is \emph{not} rotation-invariant.

	Next, the retail item is extracted from the image.
	Canny edge detector~\cite{ref_canny} performs non-maximal suppression and hysteresis thresholding to generate a binary image containing the edges, which is then subjected to morphological closing using a $7\times7$ rectangular structuring element to fill out holes and gaps.
	Objects within the image are then identified using the Border Following algorithm of \cite{ref_borderfollowing}.
	Apart from the retail item to be identified, reflection of the lighting system and empty background portions (which emerge after rotation correction) are also identified as \emph{objects} by the algorithm (see Fig.~\ref{fig:prep_steps}).
	To counter this, a minimum bounding rectangle is approximated around each identified object and the rectangle with the longest diagonal is selected.
	The whole image is then cropped to this rectangle.

	The image is then sharpened using the standard Sobel kernel.
	We then apply Histogram Equalization on the \emph{luma} component of the image to enhance contrast in the image and make the constituent object's features more prominent.
	This acts on the image by first calculating a histogram of all pixel values in the image and then normalizing the histogram so that sum of histogram bins equals $255$.
	Integral of this histogram is computed as:
	$\hat{H}_i=\sum_{0 \leq j < i } {H(j)} \label{eq:hist}$.
	Using $\hat{H}$ as a lookup table, image intensities are transformed to their new values.
	Image is then resized to $150\times150$ while keeping aspect ratio of the constituent retail item consistent.
	A step-by-step example of the outputs from our pre-processing pipeline is provided in Section~\ref{ssec:prep_sample}.

	\subsection{Neural Network Architecture}\label{ssec:cnn}
	\paragraph{Convolutional Layers}
	Recent breakthroughs in object identification and pattern recognition primarily revolved around the use of Convolutional Neural Networks (CNNs)~\cite{ref_krizhevsky, ref_inception}, which are a class of Deep Neural Networks consisting of one or more \emph{convolutional} layers.
	A convolution operation involves applying a kernel element throughout a spatially arranged input (image) at various locations.
	Each unit of the resulting score map is connected to the input pixels through kernel weights.
	These weights are optimized for the task at hand via backpropagation during training.
	Since each output unit is related to only those pixels of the input that participated in score calculation through the kernel weights, the number of parameters is significantly small as compared to the standard \emph{fully-connected} layers, making convolutional layers highly efficient, less prone (via an inductive bias) to the horrors of overfitting, and suitable for feature extraction.
	Usually more than one kernel is applied at each layer, extracting multiple features in parallel.
	Output from each of these layers is subjected to a non-linear function and connected one after the other in a cascaded fashion, going from fine image details in the initial layers to complete objects and regions in subsequent layers~\cite{ref_visualizenns}.
	If $l$ is the layer in question and $k_l$ represents a kernel element on this layer, then $y_l$ is the layer output obtained by convolving this kernel with the input and adding a learned bias $b_l$.
	If $C_{l-1}$ is the number of channels/feature-maps in the input to this layer and $f_l(\dots)$ represents a non-linearity, applying a single convolution operation on an input, combining respective feature-maps to produce $n$-th feature-map of the output, is expressed as: $y_l^n = f(\sum_{m=1}^{C_{l-1}} {k_l^{n,m} * y_{l-1}^m + b_l^m}) \label{eq_conv}$.

	\paragraph{Fully-Connected Layers}
	These layers act in a similar fashion to convolutional layers except that each element in the input is connected, via a linear transformation, to each element in the output.
	While convolutional layers handle the feature extraction task, fully-connected layers combine the features and perform the discriminative \emph{classification} task.
	Expression for a fully-connected layer with weight matrix $W_l$ for the connections is obtained as: $y_l = f(W_l * y_{l-1} + b_l) \label{eq_fc}$.

	\paragraph{Activation Functions}
	Non-linearities applied to multiple layers enable the network to learn increasingly complex functions.
	Rectified Linear Units (ReLUs)~\cite{ref_relu}, defined as: $\mathrm{ReLU}(y_l) = \mathrm{max}(0, y_l)$, are non-symmetric activation functions that were found to achieve better results than previously used candidates like Sigmoid and Hyperbolic-Tangent, while speeding up training~\cite{ref_krizhevsky}.
	But the truncation of negative inputs can impair gradient flow, for which a variant was proposed in \cite{ref_maas}.
	This modified non-linearity has a \textit{leakyness} parameter $\alpha$ to allow scaled-down negative values, hence the name Leaky ReLU: $\mathrm{LeakyReLU}(y_l) = \mathrm{max}(0, y_l) + \alpha \mathrm{min}(0, y_l)$.
	Generalization of this concept, called Parametric ReLU, was proposed in \cite{ref_he_preluinit}, where the leakyness hyper-parameter is converted into a channel-wise learnable parameter for each layer's output.
	This modification, although increases the dimensionality of the search space, adaptively learns the parameter along with the model.

	The final classification layer is usually subjected to Softmax activation, which converts the outputs into pseudo-probabilities.
	Result of this function depends on all output units of the layer, in contrast to other activations, which do not depend on output units other than the ones they act on.
	If $z^{(i)}=y^{(i)}_l$ represents the unactivated output of the $i$-th node in layer $l$ of size $K$, then the standard softmax function is defined as:
	\begin{equation}\label{eq_softmax}
	\sigma_l (y_l) = \left[ \frac{e^{z^{(1)}}}{\sum_{j=1}^K e^{z^{(j)}}}, \dots, \frac{e^{z^{(K)}}}{\sum_{j=1}^K e^{z^{(j)}}} \right]^T
	\end{equation}

	\paragraph{Max-Pooling}
	Pooling operation combines spatially nearby features in a feature-map to produce a sub-sampled version.
	This removes redundant features while retaining more significant ones in the neighborhood, enabling a massive reduction in computational requirements for further operations.
	Furthermore, this makes subsequent feature maps more robust to slight changes in position of a feature in the input, also called \textit{local translation invariance}.
	Max-Pooling does this by convolving the input with a kernel (of defined size which specifies spatial extent of neighborhood) of a $\mathrm{max}(\dots)$ operation.

	\paragraph{Weight Initialization}
	Initialization of network parameters is of prime importance to control how supervision signal propagates through the network.
	If care is not taken to retain the variance of the forward and back-propagated signals, they may explode or vanish.
	We therefore utilize the initialization scheme for rectifier-based deep neural networks proposed in \cite{ref_he_preluinit} (values drawn from a Gaussian distribution with a standard deviation of $\sqrt{2/N}$, where $N$ is the number of incoming nodes to a single neuron) for weights, while biases are initialized to $0.1$.

	\paragraph{Dropout and Batch Normalization}
	To reduce risk of overfitting, we use Dropout~\cite{ref_dropout} which randomly drops units in a network layer at a predefined rate.
	Intuitively, this forces the network to produce desired results even in the absence of certain information, allowing it to depend less on only a certain subset of inputs.
	We also make use of Batch Normalization~\cite{ref_bn} which has been shown to enable faster training and better generalizability by helping signal propagate through the network and making the optimization landscape smoother~\cite{ref_bn_how}.

	The light-weight architecture devised for this system is shown in Table~\ref{table:network}.
	Each convolutional layer is followed by Batch Normalization and PReLU activation, while each of the first two fully-connected layers is followed by PReLU activation and Dropout of $0.1$.
	Output from the final layer is subjected to Softmax activation.

	\begin{table}[htbp]
		\caption{
			\raggedright
			\textbf{CNN Architecture}.
			The feedforward convolutional neural network architecture, used as the object identification model in our setup, is detailed below.
			Each layer is detailed in sequence.
			Each convolutional layer (Conv.) is followed by a Batch-Norm layer and PReLU activation.
			Each fully-connected layer (F.C.) is followed by PReLU activation and dropout of $0.1$.
		}

		\begin{tabular}{ccccc}
			\textbf{Type} & \textbf{Filter Size} & \textbf{Stride} & \textbf{Filters/Units} & \textbf{In-Shape} \\
			\hline
			Conv. & $3\times3$ & $1\times1$ & $8$ & $150\times150\times3$ \\
			Max-Pool & $4\times4$ & $4\times4$ & - & $148\times148\times8$ \\
			Conv. & $3\times3$ & $1\times1$ & $8$ & $37\times37\times8$ \\
			Max-Pool & $2\times2$ & $2\times2$ & - & $35\times35\times8$ \\
			F.C. & - & - & $512$ & $2312$ \\
			F.C. & - & - & $256$ & $512$ \\
			F.C. & - & - & $100$ & $256$ \\
		\end{tabular}
		\label{table:network}
	\end{table}

	\begin{figure*}[tbp]
    \centering
        \includegraphics[clip, trim=13cm 0cm 5cm 0cm, width=\textwidth]{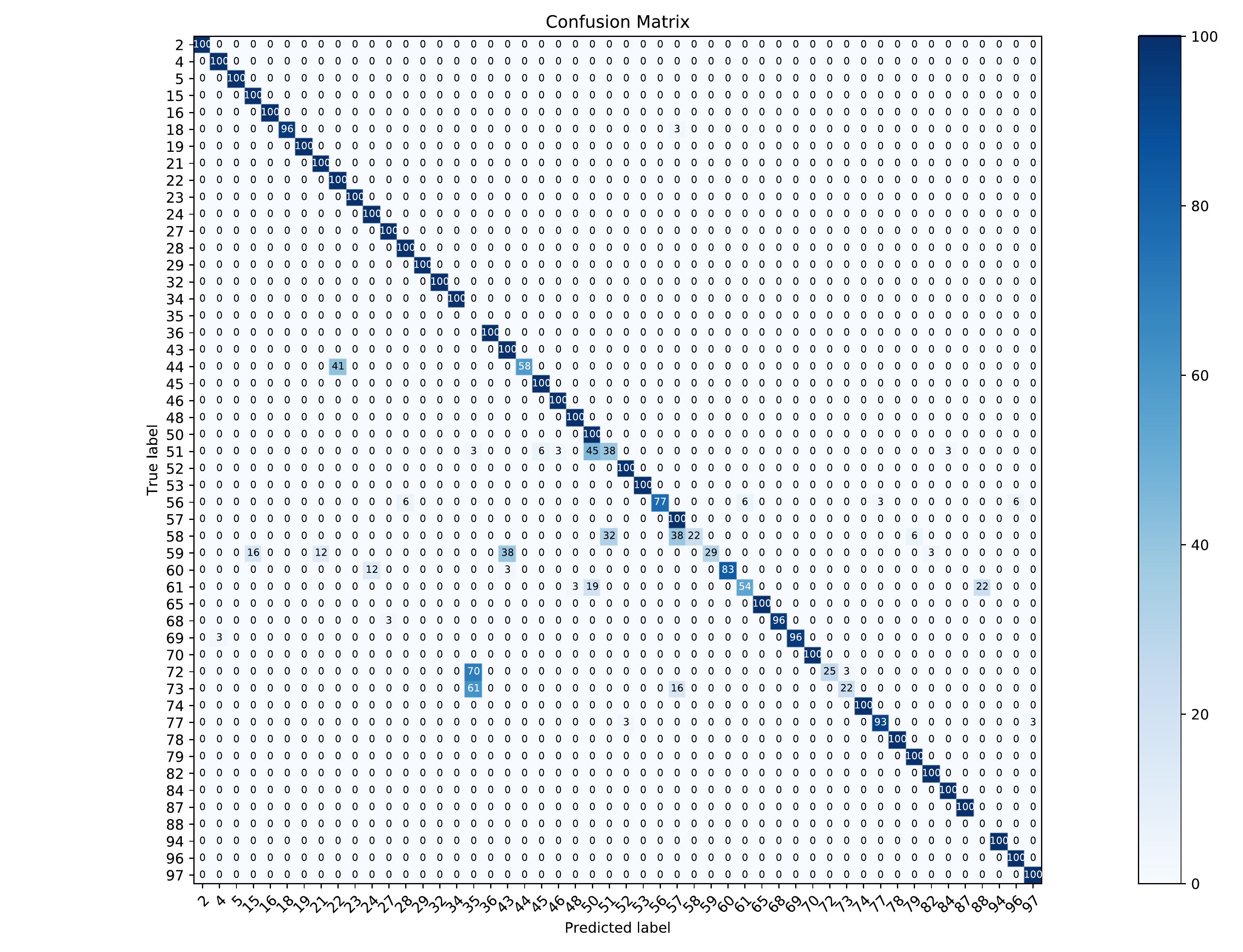}
    \caption{
			\textbf{Confusion Matrix of Selected Items}.
			To conserve space, we show only those items that are \emph{at least once} confused with other items by the neural network.
			Names of actual retail items are replaced with labels $1$-$100$ for brevity.
			Each cell in the matrix denotes frequency in \%.
		}
    \label{fig:confusemat}
	\end{figure*}

	\subsection{Billing}
	When the retail item is identified, it is added to the shopping cart with its price, visible on the GUI.
	Once all items have been scanned and added to the shopping cart, user checks out and a receipt is collected from the printer.


	\section{Experimentation}\label{sec:experiment}

	\subsection{Data Curation}
	To train our neural network, we curated a dataset of local retail items, for which we visited Hyperstar Pakistan (now Carrefour Pakistan) and gathered 100 different products.
	Each product was pictured under the webcam inside the hood, at different \emph{logical} orientations where the important distinguishable features of the item are evident.
	This rules out empty sides and sides containing only general information like ingredients and nutrition content.
	The underlying hypothesis is that the system should be able to identify items by looking at its characteristic features, like a human observer would do.
	A total of 31,000 images (divided equally among the retail items) were acquired.
	We performed a 65-25-10 split for training, validation and testing.

	\subsection{Training of Neural Network}
	For training as well as validation, $32$ images form a mini-batch over which an optimization step is performed.
	This consists of $16$ images straight from the pre-processed dataset, with their randomly rotated versions from the set of angles $\{90, 180, 270\}$.
	During training, we optimize the Categorical Cross-Entropy Loss, which can be stated as:

	\begin{equation}
	\mathrm{CE}(y, \hat{y}) = \sum_{i \in units} \sum_{j \in classes} y_{i, j} \log(\hat{y}_{i, j})
	\end{equation}

	\noindent where $\hat{y}$ is the output of the network for a certain input and $y$ is the corresponding one-hot encoded ground truth.
	AMSGrad variant of Adam Optimizer~\cite{ref_adam_ams}, is used with $\beta_1=0.9$ and $\beta_2=0.999$ as the exponential decay rates for moment estimates.
	Weights and biases are penalized with an $L_2$ penalty (weight decay) of $0.01$, to encourage diffused values.
	Maximum training epochs were set to $100$, starting with a learning rate of $0.001$.
	Learning rate is scheduled to reduce after each epoch, with a decay rate of $0.96$ till $20$-th epoch, after which the decay rate is reduced to $0.75$.
	Additionally, if validation loss plateaus the learning rate is reduced to $1/10$-th of its instantaneous value.
	We used the resources at Google Colaboratory, a free cloud service based on Jupyter Notebook, to train our neural network.
	The light-weight network reached a training accuracy of $94.76\%$ and a validation accuracy of $95.24\%$.

	\subsection{Results and Discussion}
	At the present stage, system design involves a few working assumptions:
	\begin{itemize}
		\item Each retail item is fed to the system turn by turn, i.e. at any point no two objects appear within the field-of-view of the webcam during image acquisition.
		\item Retail items are neither too small for the camera to acquire characteristic features, nor too large to fit inside the wooden hood.
		\item Conveyor is stopped whenever a frame is to be extracted from the continuous webcam feed, to avoid motion blur.
	\end{itemize}

	Under these assumptions, our model yields an overall testing accuracy of $91.7\%$.
	The deficiency in achieving a perfect score can be explained by observing the object-wise performance in Fig.~\ref{fig:confusemat}.

\begin{figure}[!t]
		\centering
		\subfloat[Item 21]{\includegraphics[width=0.28\linewidth]{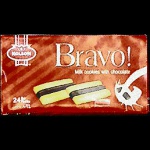}
			\label{fig:worstcase_21}}
		\hfil
		\subfloat[Item 87]{\includegraphics[width=0.28\linewidth]{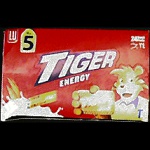}
			\label{fig:worstcase_87}}
		\hfil
		\subfloat[Item 73]{\includegraphics[width=0.28\linewidth]{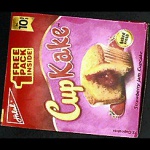}
			\label{fig:worstcase_73}}
		\hfil
		\subfloat[Item 74]{\includegraphics[width=0.28\linewidth]{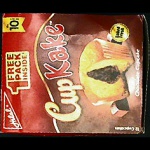}
			\label{fig:worstcase_74}}
		\hfil
		\subfloat[Item 2]{\includegraphics[width=0.28\linewidth]{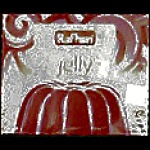}
			\label{fig:worstcase_2}}
		\hfil
		\subfloat[Item 36]{\includegraphics[width=0.28\linewidth]{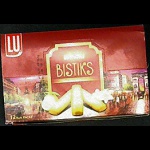}
			\label{fig:worstcase_36}}
		\caption{
			\textbf{Worst Performance Cases}: Misclassification rate of~\ref{fig:worstcase_87} with~\ref{fig:worstcase_2} is 38.71\%, \ref{fig:worstcase_21} with~\ref{fig:worstcase_74} is 83.87\%, and \ref{fig:worstcase_36} with~\ref{fig:worstcase_73} is 41.94\%, while that of~\ref{fig:worstcase_73} with~\ref{fig:worstcase_36} is 70.97\%.
			(Images are pre-processed before being fed to the neural network.)
		}
		\label{fig:worstcase}
	\end{figure}

	The worst mis-classifications are of Item 21 with 74, 87 with 2, 73 with 36, and 74 with 36 (displayed in Fig.~\ref{fig:worstcase}).
	In some of these cases, the items have subtle differences in their processed appearance that is being fed to the neural network.
	Note that, for example, although item 87 is primarily confused with item 2, item 2 itself is never confused with any other item, i.e. the network appears to be biased towards certain classes.
	This begs further experimentation to improve discrimination, a possible solution could be the use of weighted cross-entropy loss function where classes that the network fails to recognize are penalized more.
	Along with this, the glare from glossy and reflective surfaces of objects makes it an even more tedious task.
	Better camera and a more diffused illumination system can provide a better quality input that may improve the discriminative power of the network.

\begin{figure}[!t]
		\centering
		\subfloat[Item 95]{\includegraphics[width=0.28\linewidth]{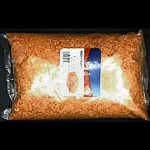}
			\label{fig:bestcase_95}}
		\hfil
		\subfloat[Item 38]{\includegraphics[width=0.28\linewidth]{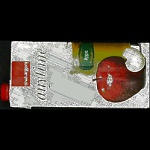}
			\label{fig:bestcase_38}}
		\hfil
		\subfloat[Item 84]{\includegraphics[width=0.28\linewidth]{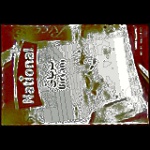}
			\label{fig:bestcase_84}}
		\hfil
		\subfloat[Item 96]{\includegraphics[width=0.28\linewidth]{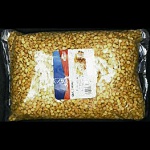}
			\label{fig:bestcase_96}}
		\hfil
		\subfloat[Item 42]{\includegraphics[width=0.28\linewidth]{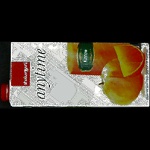}
			\label{fig:bestcase_42}}
		\hfil
		\subfloat[Item 86]{\includegraphics[width=0.28\linewidth]{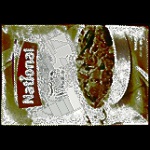}
			\label{fig:bestcase_86}}
		\caption{
			\textbf{Best Performance Cases}: \ref{fig:bestcase_95} and~\ref{fig:bestcase_96}, \ref{fig:bestcase_84} and~\ref{fig:bestcase_86}, and~\ref{fig:bestcase_38} and~\ref{fig:bestcase_42} appear quite similar to each other but are never misclassified by the network.
			(Images are pre-processed before being fed to the neural network.)
		}
		\label{fig:bestcase}
	\end{figure}

	On the contrary, the network does not fail in recognizing some similar looking items that one would expect it to, such as items 95 and 96, items 84, 85 and 86, and 38 and 42 (displayed in Fig.~\ref{fig:bestcase}).
	This may be accredited to the pre-processing pipeline that makes it easier to extract more discriminative features which would otherwise be suppressed by the poor camera quality and surface glare.

	\section{Conclusion and Future Work}\label{sec:conclude}
	We propose a computer vision-based retail checkout system which makes use of image processing techniques coupled with a convolutional neural network to identify an retail items beneath a webcam.
	One of the key goals kept in view while designing this system, other than improving the current prevailing checkout practices and demonstration of an innovative method, was cost-effectiveness.
	Implementation costs and resources at hand were kept in consideration, therefore the system makes use of inexpensive components, while neural network was trained on a free web service.

	Working with a webcam has its downsides, in the form of image blurring due to motion.
	A camera with a higher shutter speed will enable capturing objects without the hassle of stopping the conveyor every time.
	Combined with a better object detection technique, such as YOLO~\cite{ref_yolo}, the method can detect and separate multiple objects in real time, and feed them separately to our network for inference.
	The image database is experimental, and will be expanded to include more objects, begging for a better technique to identify objects than brute force classification.
	An important aspect that we envisage in the final form of this setup is the ability to incorporate new items without much re-training.
	For this, emphasis is needed to study the feature extraction portion of the network so that it is able to extract more rich and generic features which could be conveniently modified for the task at hand via fine-tuning.

	\nocite{*}
	\bibliographystyle{IEEEtran}
	\bibliography{bibliography/references}

	\appendices

	\section{Example: Pre-processing}\label{ssec:prep_sample}

\begin{figure*}[!ht]
		\centering
		\subfloat[Input image]{\includegraphics[width=0.28\linewidth]{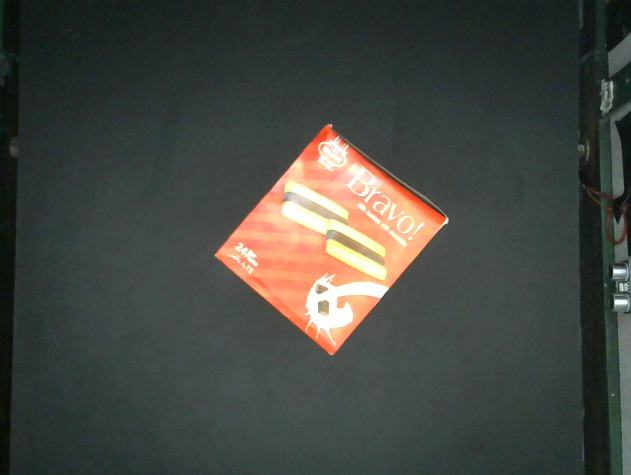}
			\label{fig:img_inp}}
		\hfil
		\subfloat[Rotation corrected (after cropping)]{\includegraphics[width=0.28\linewidth]{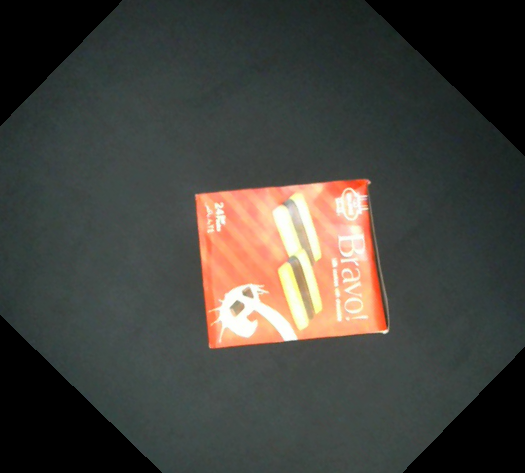}
			\label{fig:img_rot}}
		\hfil
		\subfloat[Edge detection binary mask]{\includegraphics[width=0.28\linewidth]{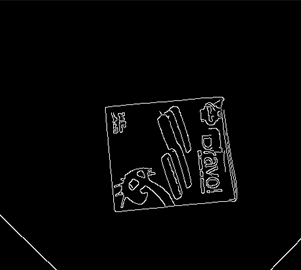}
			\label{fig:img_edgedet}}

		\subfloat[Morphological closing over \ref{fig:img_edgedet}]{\includegraphics[width=0.28\linewidth]{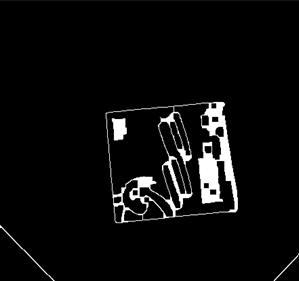}
			\label{fig:img_close}}
		\hfil
		\subfloat[Bounding boxes from \ref{fig:img_close}]{\includegraphics[width=0.28\linewidth]{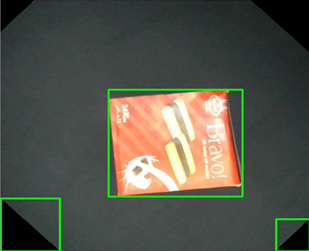}
			\label{fig:img_bbox}}
		\hfil
		\subfloat[Segmented and cropped]{\includegraphics[width=0.28\linewidth]{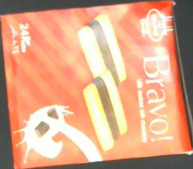}
			\label{fig:img_segcrop}}

		\subfloat[Sharpened]{\includegraphics[width=0.28\linewidth]{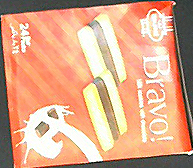}
			\label{fig:img_sharp}}
		\hfil
		\subfloat[Histogram equalized]{\includegraphics[width=0.28\linewidth]{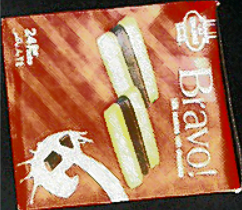}
			\label{fig:img_histeq}}
		\hfil
		\subfloat[Zero-padded and resized]{\includegraphics[width=0.28\linewidth]{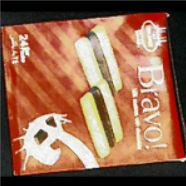}
			\label{fig:img_out}}

		\caption{
			\textbf{Pre-processing Steps}.
			Figure shows the outputs generated, in sequence, by our pre-processing pipeline of Fig.~\ref{fig:preprocessing}.
			Note that features of the output image (\ref{fig:img_out}) are significantly more visible than in the input image (\ref{fig:img_inp}), which aid in the classification task.
			We use a black conveyor belt with a matte finish to make the input object prominent and to avoid glare produced due to the LEDs.
			First, the input image is cropped (via a predefined cropping) to remove regions outside the conveyor belt, such as the conveyor support structure visible in \ref{fig:img_inp}.
			The object rotation is corrected as in \ref{fig:img_rot}, where black (zero intensity) regions are visible.
			Edge detection attempts to outline borders of regions of interest (\ref{fig:img_edgedet}).
			We additionally perform morphological closing to make the object of interest more salient (\ref{fig:img_close}).
			Bounding boxes are generated on the resulting image, shown in \ref{fig:img_bbox}, where we select the box with the longest diagonal length and crop to it (\ref{fig:img_segcrop}).
			The image features are enhanced by sharpening (\ref{fig:img_sharp}) and histogram equalization (\ref{fig:img_histeq}).
			Finally, to standardize the input dimensions to the neural network, the image is padded with zeros to yield a square matrix and resized to $150\times150$.
		}
		\label{fig:prep_steps}
	\end{figure*}

	Fig.~\ref{fig:prep_steps} shows how pre-processing impacts the images at each step.
	As mentioned in Section~\ref{ssec:preprocessing}, the pre-processing pipeline is designed to standardize the input to the neural network while enhancing the discriminative features of the image.

\end{document}